\documentclass{article}

\PassOptionsToPackage{numbers,sort&compress}{natbib}

\usepackage[preprint]{neurips_2026}

\newif\ifanon
\anonfalse     

\usepackage[utf8]{inputenc}
\usepackage[T1]{fontenc}
\usepackage{hyperref}
\usepackage{url}
\usepackage{booktabs}
\usepackage{amsmath}
\usepackage{amssymb}
\usepackage{microtype}
\usepackage{xcolor}

\title{The LLM Proposes, the Executive Disposes:\\
A Self-Verifying Agent Instrument that Dissociates\\
Commitment Drift from Binding Drift in Long-Horizon Agents}

\ifanon
  \author{Anonymous Author(s)\\Affiliation\\\texttt{email}}
\else
  \author{%
    Mohsen Arjmandi\\
    Independent researcher\\
    \texttt{mohsen.arjmandi@gmail.com}\\
    \texttt{\href{https://github.com/arjmandi/ARG}{github.com/arjmandi/ARG} (tag \texttt{v1.0-paper})}%
  }
\fi

\begin{document}
\maketitle

\begin{abstract}
How do you verify a long-horizon agent when its own state and self-reports are
exactly what you cannot trust? We present an agent instrument built so that
verification is structural rather than post-hoc. A deterministic \emph{Executive}
owns all belief; a language model may only file typed proposals, and a claim is
admitted only when a prediction \emph{pre-registered before acting} is matched
against observation by code. Two properties make the instrument a verifier of its
own science, not just of the agent: every run \emph{invalidates itself} when
per-organ write-error, render-size, or salted-canary-echo floors are breached
(four of the first eight architecture runs were invalidated, each localizing a
real defect); and a render-invisible \emph{shadow reference} compiles the plan the
full system would have committed \emph{in every ablation cell}, so drift metrics
are defined even where the mechanism under test has been removed. Using this
instrument we report a clean, single-variable result on a failure every
long-horizon agent suffers: ablating the commitment mechanism flips
goal-abandonment from $0.00$ to $1.00$ while binding error stays flat at $0.00$
(three seeds per cell, up to 394 reference beats per run, every run gated valid).
The binding channel, by contrast, does not reappear as per-beat drift when its
repair is ablated --- because binding is code-owned, the failure class is
\emph{structurally absorbed}, its only residue appearing one layer upstream as a
collapse in hypothesis formation. We report these under full disclosure that task
efficacy is null (zero level completions across 52 gated runs on ARC-AGI-3),
pre-registered as a structural defeater. The contribution is a verification
methodology for agent development and the drift decomposition it makes measurable.
\end{abstract}

\section{Introduction}

An agent that acts over hundreds or thousands of steps drifts off its own
objective. The phenomenon --- \emph{goal drift} --- is now measured and
benchmarked as a first-class failure of language-model agents
\citep{arike2025goaldrift,menon2026inherited}. But the way we \emph{verify} it is
weak in two ways that matter for reliable agent development. First, most
evaluations trust the agent's own state and self-reports; an agent that says
``done'' is scored as done. Second, drift is reported as an aggregate score that
cannot say \emph{which} failure occurred, or whether a given run was even valid to
attribute a number to. This paper takes the workshop's question --- \emph{who
verifies the agent?} --- literally, and answers it by construction.

We describe an instrument in which verification is not a downstream check but the
substrate. Belief is earned, not asserted: a claim enters the agent's state only
when a prediction committed to a log \emph{before} acting is matched against the
environment's response by deterministic code. The agent cannot author a match it
did not pre-commit to, and ``done'' is never an event. On top of this we add two
verifiers of the \emph{measurement}: a run-validity gate that makes a run
invalidate itself when the machinery is misbehaving, and a shadow instrument that
defines the drift metrics in every experimental cell, including the cells where
the mechanism under test has been deleted.

That instrument buys a result. ``Drift'' is not one failure but at least two,
with different signatures and different repairs:
\begin{itemize}
\item \textbf{Binding drift} --- the intention and its referents are both present,
but the join between ``the goal'' and ``the object it is about'' lives only in
transformer attention, which loses bindings under distance and low salience (the
positional / spaced-evidence regime; \citealp{longpibench2024,liu2023lost}). The
repair is \emph{positional}.
\item \textbf{Commitment drift} --- the intention is simply absent from later
contexts, and error compounds even when every context is short and the goal
maximally salient (the goal-abandonment regime measured in the agent literature;
\citealp{arike2025goaldrift,menon2026inherited}). The repair is \emph{not}
positional: an external commitment store executed by code.
\end{itemize}
The binding component is not merely hypothesized: prior work on a predecessor
system~\ifanon\else\citep{arjmandi2026sensi}\fi{} diagnosed a
\emph{self-consistent hallucination cascade} in which perception-layer errors
propagate into internally coherent but factually wrong world models. What that work names a perception-layer cascade is,
in the present taxonomy, binding loss --- a referent grounded in the model's own
prior text rather than in observation. Our instrument externalizes a repair for
each failure as an independent switch, so the two can be turned off, and measured,
separately.

\paragraph{Contributions.}
\begin{enumerate}
\item \textbf{A self-verifying agent instrument} (\S\ref{sec:arch}--\S\ref{sec:verif}):
a deterministic control plane that owns belief; runs that invalidate themselves on
per-organ error, render-size, and canary-echo floors; and a render-invisible
shadow reference that defines drift in every ablation cell against a byte-identity
guarantee.
\item \textbf{A causal dissociation} (\S\ref{sec:results}): ablating the
commitment mechanism drives goal-abandonment to the ceiling ($0.00\!\to\!1.00$)
with binding error flat, while ablating the binding repair changes nothing
per-beat --- evidence that ``drift'' should be decomposed before it is repaired.
\item \textbf{The null, reported first} (\S\ref{sec:null}): zero task
completions across 52 gated runs, pre-registered as a defeater --- and a
demonstration that the mechanism results hold independently of it.
\end{enumerate}

\section{The verification architecture}
\label{sec:arch}

\paragraph{The LLM proposes, the Executive disposes.} No model output mutates state
and no model reads a raw dump. Three model ``organs'' --- an Observer that
interprets machine-computed change-sets, a Surveyor that files goals / hypotheses /
experiments, and an Actuator that emits actions --- may only submit typed proposals
in closed vocabularies through admission gates. A deterministic \textbf{Executive}
(differ, matcher, validator, compiler, renderer, test-evaluator) owns every state
transition. All stores are append-only; status is \emph{computed}, never
overwritten; every consequence receipt keys to the exact per-frame anchor under
which it was observed, so identity is replayable data rather than a destructive
merge.

\paragraph{Belief is earned by verification.} A referent climbs tiers, each
transition machine-checked: named-only text is not yet a referent; a
salience-blind extraction from raw observation anchors it; a logged action receipt
engages it; and it becomes \emph{characterized} only when referenced by a rule
that is \textsc{tested} --- i.e., whose predictions, committed to the log
\emph{before} acting, were matched against the deterministic observation diff by
the Executive. This is verification in the strict sense: the model cannot author a
match it did not pre-commit to, achievement is a fired predicate over logged
events (``an LLM saying done is not an event''), and demotion never deletes the
audit chain.

\paragraph{The two switches.} The binding repair \textbf{J} is the
goal$\leftrightarrow$referent \emph{join}: a persistent structure that
canonicalizes referents from consequence-tested evidence and re-surfaces every
active goal pre-joined with its referents at fixed prompt positions each turn, so
no consumer joins facts across distant prompt regions. The commitment repair
\textbf{A} is the external commitment store executed by code: a compiled plan the
Actuator consults independent of model attention. Each is a single environment
flag (\texttt{ARG\_JOIN=0}, \texttt{ARG\_AGENDA=0}); all other scaffolding is
identical across cells, so a cell-to-cell delta cannot be attributed to generic
scaffolding or to call count. Nothing above a small, declarative environment
adapter is task-specific --- the property that makes the ``unseen game'' result
(\S\ref{sec:null}) meaningful, enforced at release by an exemplar scrub.

\section{Verifying the measurement itself}
\label{sec:verif}

A dissociation is only as trustworthy as the instrument that measures it. Three
mechanisms verify the measurement before any belief-level claim is permitted.

\paragraph{Runs that invalidate themselves.} A run is \textsc{invalid} for
attribution --- excluded entirely --- if any per-organ write-error rate exceeds
$0.25$ (typed rejection rows over total ops), if any rendered view exceeds the
hard token ceiling per beat \emph{or} per call, or if salted-canary echo rates per
prompt zone fall below floor. Canaries are quarantined synthetic rows injected into
every zone; per-zone echo separates transport defects from consumption rot before
any belief-level diagnosis. This is not decorative: four of the first eight
architecture runs were invalidated by these floors, and each invalidation
localized a real contract defect. Every number in \S\ref{sec:results} is from a
valid run and is replayable from an append-only log.

\paragraph{Drift defined in every cell.} Goal drift is deviation from a plan --- but
in commitment-ablated cells there is no plan to deviate from. The instrument closes
this by \emph{shadow-compiling}, in every cell, the plan the full system would have
committed, without rendering or executing it: a pure read over the same store. A
byte-identity test proves the shadow path cannot leak into behavior (rendered bytes
are identical with the shadow on or off). This yields a positive reference-beat
count --- and therefore \emph{defined} bind and abandon scores --- even in cells
that have no live plan at all. Without it, the commitment half of the dissociation
would rest on an undefined metric.

\paragraph{Operational drift taxonomy.} For each reference beat with a plan step
$(\text{target}, \text{action})$, the beat is scored \textbf{bind} if it took the
plan's action with an aim but landed on a referent $\neq$ the plan's target;
\textbf{aligned} if it faithfully executed a live step; \textbf{abandon} if it
neither followed the plan's action nor hit the plan's deliberate target. Untargeted
actions carry no aim parameter and so present no binding seam --- they can abandon
but cannot bind-miss, which is why bind score is structurally near-zero wherever
plans are untargeted (relevant in \S\ref{sec:absorb}).

\paragraph{Compute parity.} The full system runs against a bare backbone at
$\rho_{\text{calls}}=0.15$ and $\rho_{\text{tokens}}=0.38$ --- it spends
\emph{less} inference, not more --- so no reported effect can be attributed to
extra inference spend. Efficiency figures are reported only for completed levels;
a zero-completion cell prints ``0 completions'' and no efficiency number, ever.

\section{Results: a dissociation the instrument makes visible}
\label{sec:results}

Campaign to date: 52 runs, ${\sim}17$M tokens, two protocol days, on ARC-AGI-3
interactive games~\citep{arcagi3_2026}, a mid-tier backbone unless noted,
200--400-action horizons. Every figure is from a valid run.

\subsection{The commitment-drift dissociation}

At protocol seed count on one game (three seeds per cell, every cell valid):

\begin{table}[h]
\centering
\begin{tabular}{lccccc}
\toprule
Cell & binding repair \textbf{J} & commitment repair \textbf{A} & GDS-bind & GDS-abandon & ref.\ beats/run \\
\midrule
FULL       & on  & on           & $0.00$ & $0.00$ & up to 96 \\
J0         & off & on           & $0.00$ & $0.00$ & up to 72 \\
\textbf{A0}& on  & \textbf{off} & $\mathbf{0.00}$ & $\mathbf{1.00}$ & up to 394 \\
J0A0       & off & off          & --- & --- & (undefined; \S\ref{sec:absorb}) \\
\bottomrule
\end{tabular}
\end{table}

\noindent Killing the commitment store flips goal-abandonment $0.00\!\to\!1.00$
with binding error flat at $0.00$, wherever the metric is defined (abandon defined
on $n{=}2$ of the 3 A0 seeds; the third drew no feedstock and is undefined, not
zero). The effect is single-variable: FULL and J0 both hold abandonment at $0.00$;
only removing A moves it, and it moves to the ceiling, against up to 394
shadow-compiled reference beats in a single run. Deprived only of its external
commitment store, the agent abandons the very plan it would otherwise have followed
on essentially every beat, while its binding behavior is unchanged.

\subsection{Binding drift is structurally absorbed}
\label{sec:absorb}

The binding half did not behave as a symmetric double dissociation would predict,
and the reason is itself a result. Bind score is $0.00$ in \emph{every} cell above,
including the join-killed cells. We confirmed this is absorption, not an idiosyncrasy
of one game, on a second, targeted-action-rich game in a paired \emph{engaged}
comparison:

\begin{table}[h]
\centering
\begin{tabular}{lcc}
\toprule
Metric (game 2, paired engaged) & FULL & J0 (join killed) \\
\midrule
GDS-bind        & $0.00$  & $0.00$  \\
GDS-abandon     & $0.00$  & $0.00$  \\
Grounding rate  & $1.00$  & $1.00$  \\
Goal adherence  & $0.964$ & $0.964$ \\
\bottomrule
\end{tabular}
\end{table}

\noindent Killing the model-facing join changes nothing per-beat. The
binding-failure class the join was built to guard cannot open here, because
aiming, containment checking, and receipt attribution are Executive-owned code:
an aimed emission's landing is decided by geometry (which referent's anchor cells
contain the emitted coordinates), and a wrong landing is caught and its receipt
suppressed \emph{before} it can corrupt belief. The join's contribution has moved
from ``prevent per-beat binding loss'' to structure, where it does not print as
drift. One integrity note makes the $0.00$s credible rather than suspicious: an
early containment bug attributed a click to the \emph{first} enclosing referent by
id rather than the most specific one --- 96 receipts mis-attributed in a single run,
silently starving the aimed rule and \emph{printing fake drift}. The instrument's
own decomposition surfaced it; the fix (attribute the smallest containing referent)
was pinned by a fixture the same day. The clean $0.00$s are the corrected reading.

\subsection{The residue is upstream}

If the join's load is structural, removing it should show up somewhere. Not
per-beat, but one layer earlier, in whether effect-linked hypotheses form at all.
Across the factorial, hypothesis-formation rate runs
$\text{FULL } 3/3 = \text{J0 } 3/3 > \text{A0 } 2/3 > \text{J0A0 } 0/4$. Removing
both mechanisms floors the system \emph{upstream} of drift: with neither the join
nor the agenda context available, effect-linked hypotheses essentially never form
($0$ of $4$ seeds). This is why the double-kill cell is undefined for drift in
\S\ref{sec:results} --- the system never gets far enough to have a plan to drift
from --- and it relocates the binding repair's measurable effect to hypothesis
\emph{feedstock}, one step removed from binding execution.

\section{The efficacy null, small-model floor, and generality}
\label{sec:null}

\paragraph{Task efficacy is null, stated first.} Across all 52 runs and every cell
--- including every baseline --- there are zero level completions and zero score.
We pre-registered a $\sim$0-wins endpoint as a \emph{structural defeater} of the
efficacy claim rather than a caveat, and we make no efficacy claim here. The null
does not touch \S\ref{sec:results}: those are measurements against internal
reference plans, defined and gated independent of whether any level is won. It does
mean the system has not yet found the mechanic that scores; the observed behavior is
a disciplined experimentalist that recognizes its own knowledge deficits, files
targeted hypotheses, runs bounded experiments with pre-registered predictions, and
demotes wrong theories with receipts --- testing ${\sim}3$ wrong object-scoped
theories per 400-action run. Baselines are at zero too; and a subsequent analysis
finds many \emph{public} games in this family solvable by non-intelligent
strategies~\citep{liew2026explore}, so we treat 0 completions on them as a genuine
gap, not one excused by difficulty.

\paragraph{A small model runs the whole loop under the gate.} A small (Haiku-class)
backbone ran the complete loop \emph{valid at write-error rate $0.0$}, engaged
throughout (2 hypotheses $\to$ 2 milestones $\to$ 48 aimed experiment beats, zero
rejected ops). This was a contract problem, not a model wall: a single lever ---
worked per-operation examples in the organ contract --- moved its write-error rate
from $0.55$--$0.71 \to 0.333 \to 0.0$. The tier gap is real (the larger model
decomposes more richly) but the floor is crossed, which bears on the economic case
for small agent models~\citep{belcak2025small}: it is testable here precisely
because verification is structural and the model's job reduces to slot-filling
machine-stated questions in a closed grammar.

\paragraph{The machinery is substrate-general.} The engine engaged on 3/3
never-before-seen games with zero game-specific code, generating curricula,
hypotheses, and milestones on all three (2/3 passed the validity gate). With no
seeded goals it generated the canonical learn-shaped curriculum verbatim from its
own typed deficits.

\section{Related work}

\textbf{Verification and verifiers.} The workshop's premise --- that agent
behavior must be checked by something other than the agent --- motivates our design:
rather than a downstream verifier scoring outputs, the deterministic side owns
belief, and the run gates itself. \textbf{Propose-and-verify} control (LLM-Modulo
planning; ``Symbolic Governor''; ``Blueprint First, Model Second'') shares the
pattern of a deterministic checker, but there the verifier checks \emph{outputs};
here it owns what the agent may treat as true. \textbf{Goal drift} is established
and benchmarked \citep{arike2025goaldrift,menon2026inherited}; we add a component
decomposition and per-cell measurement rather than a new aggregate score.
\textbf{Long-context reference / positional bias}
\citep{longpibench2024,liu2023lost} motivates the binding construct; our finding is
that once the join is code-owned the positional failure does not manifest as drift.
\textbf{Coreference caches} (LQCA-style joins; LINK-KG-style canonical-referent
caches) are prior art for the join's plumbing; the delta is that our referents
climb tiers only through Executive-verified consequence receipts, which static-text
caches lack. \textbf{Grounding} is claimed only in the causal-informational sense
relative to a formal micro-world~\citep{floridi2025categorical}; we implement the
sanctioned route of curation, verification, and tooling and claim nothing stronger.

\section{Limitations}

Efficacy is null (\S\ref{sec:null}); every claim here is a mechanism claim. The
result is a single clean isolation (commitment) plus a structural-absorption finding
(binding), not two symmetric arms: we cannot exhibit ``kill J $\Rightarrow$ bind
score rises,'' because binding is prevented by construction. The commitment
dissociation is at protocol seed count but on one game; the absorption confirmation
is one paired game; machinery generality is three games, but the full claim-bearing
protocol ($\geq 3$ games $\times \geq 3$ seeds for the dissociation itself) is not
complete, and the double-kill cell is undefined for drift, not zero. Closed
vocabularies may not span every environment; record-quantified milestones can be
gamed; consistent-but-wrong theories can pass receipts when probes lack
discriminating power. ``Grounding'' throughout denotes causal-informational
grounding relative to a formal micro-world; perceptual and social grounding are not
claimed. Run-to-run variance on identical configurations is large, so no single-run
readout is treated as meaningful --- the protocol and the gate are the only lens.

\section{Conclusion}

``Goal drift'' is not one failure. At least two mechanically distinct failures hide
inside it, with different repairs, and with an instrument that verifies its own runs
and defines drift in every cell they come apart: remove the commitment store and
nothing else, and the agent abandons its own plan on essentially every beat while
its binding stays clean; externalize binding into code and it stops printing as
per-beat drift at all. We report this with task efficacy at zero, because
the decomposition is the contribution and it holds regardless. Verifying \emph{which}
drift is happening, in every cell, is the prerequisite for repairing either --- and
it is a property you build into the agent, not one you bolt on afterward.

\ifanon
\else
\paragraph{Reproducibility.} All results are replayable from append-only logs by
the read-only instruments at \href{https://github.com/arjmandi/ARG}{github.com/arjmandi/ARG}
(tag \texttt{v1.0-paper}); the validity gate (\texttt{probe\_arg\_legibility.py}),
drift taxonomy (\texttt{probe\_arg\_metrics.py}), and release lint
(\texttt{probe\_arg\_release.py}) are pinned by hermetic tests.
\fi

\bibliographystyle{plainnat}
\bibliography{references}

\end{document}